\RequirePackage{fix-cm}

\documentclass[smallextended]{svjour3}       
\smartqed  
\usepackage{graphicx}
\usepackage{subfigure}
\usepackage{algorithmicx}
\usepackage{algpseudocode}
\usepackage{algorithm}
\usepackage{amsmath}
\usepackage{amssymb}
\usepackage[numbers]{natbib}
\usepackage{url}

\begin{document}

\title{Composing photomosaic images using clustering based evolutionary programming 
}


\author{Yaodong He      \and
        Jianfeng Zhou \and Shiu Yin Yuen
}

\author{Yaodong He      \and
        Jianfeng Zhou \and Shiu Yin Yuen
}

\institute{Yaodong He \at
              \email{yaodonghe2-c@my.cityu.edu.hk}          
           \and
           Jianfeng Zhou \at
              \email{jfzhou2-c@my.cityu.edu.hk}          
              \and
           Shiu Yin Yuen \at
              \email{kelviny.ee@cityu.edu.hk}          
            \\
              Y. He, J. Zhou and S.Y. Yuen are with the Department of Electronic Engineering, City University of Hong Kong.
}

\maketitle

\begin{abstract}
Photomosaic images are a type of images consisting of various tiny images. A complete form can be seen clearly by viewing it from a long distance. Small tiny images which replace blocks of the original image can be seen clearly by viewing it from a short distance. In the past, many algorithms have been proposed trying to automatically compose photomosaic images. Most of these algorithms are designed with greedy algorithms to match the blocks with the tiny images. To obtain a better visual sense and satisfy some commercial requirements, a constraint that a tiny image should not be repeatedly used many times is usually added. With the constraint, the photomosaic problem becomes a combinatorial optimization problem. Evolutionary algorithms imitating the process of natural selection are popular and powerful in combinatorial optimization problems. However, little work has been done on applying evolutionary algorithms to photomosaic problem. In this paper, we present an algorithm called clustering based evolutionary programming to deal with the problem. We give prior knowledge to the optimization algorithm which makes the optimization process converges faster. In our experiment, the proposed algorithm is compared with the state of the art algorithms and software. The results indicate that our algorithm performs the best.
\end{abstract}

\section{Introduction}

Photomosaic images are a type of images composed of tiny images (these tiny images are called tiles in this paper). A complete image can be recognized by viewing a photomosaic image from a long distance. The detailed vision of each tile can be seen clearly by viewing the photomosaic image from a short distance. Figure 1 is a photomosaic example generated by Mozaika software \cite{mozaika}. Silvers et al. \cite{silvers1997photomosaics} systematically describe the photomosaic problem 21 years ago. However, the visual sense of the photomosaic images is unsatisfactory at that time. In the past, generating a photomosaic image can be expensive since it requires massive calculations, thus taking lots of time. However, because of the exponential growth of computing capability, generating a photomosaic image has become much cheaper, and now the photomosaic technology is widely used in advertisements, posters and multimedia \cite{battiato2006survey, battiato2007digital}, etc.
\\

\begin{figure*}[h!]

\subfigure{
\begin{minipage}{0.48\textwidth}(a)
\centering
\includegraphics[width=5.6cm, height=4cm]{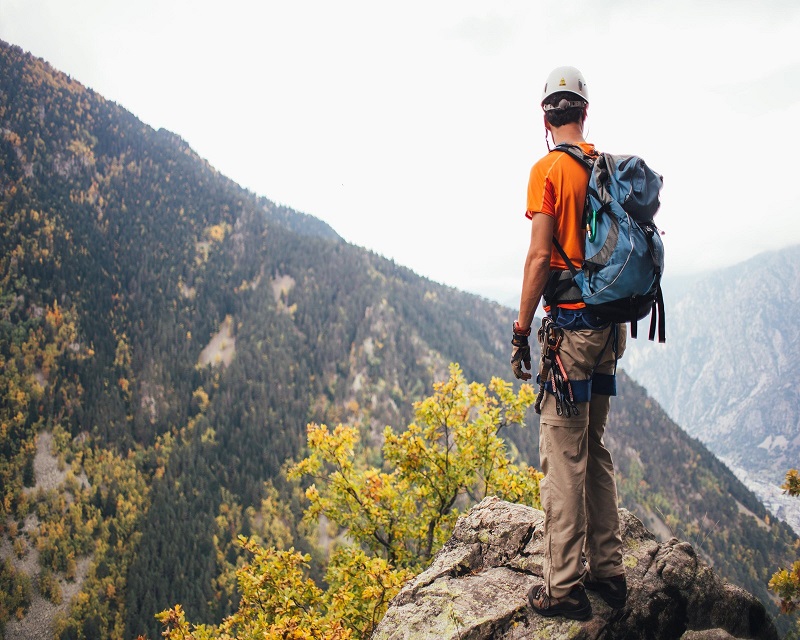}
\end{minipage}
}
\subfigure{
\begin{minipage}{0.48\textwidth}(b)
\centering
\includegraphics[width=5.6cm, height=4cm]{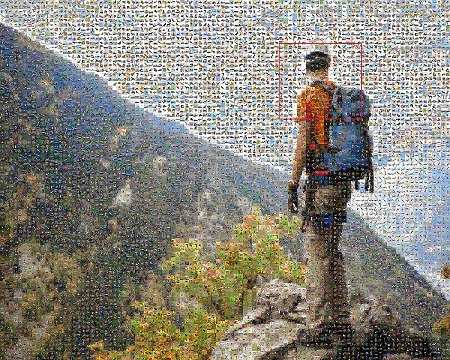}
\end{minipage}
}

\subfigure{
\begin{minipage}{0.48\textwidth}(c)
\centering
\includegraphics[width=5.6cm, height=4cm]{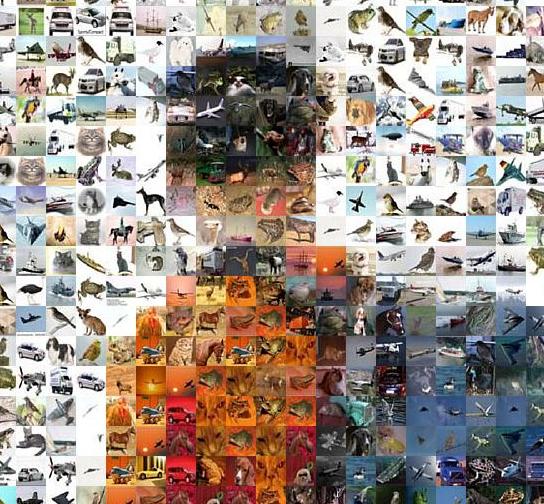}
\end{minipage}
}

\caption{(a) is the original image. (b) is the photomosaic image generated by Mozaika software. (c) is a partition of the photomosaic image. }
\end{figure*}

Usually, the photomosaic problem is solved following these steps (although different algorithms or approaches can be slightly different): 1) tiny images are collected to compose a tile database; 2) the original image is divided into blocks; 3) for each block, the intensity differences between the block and the tiles are computed to find the most suitable tile using a greedy algorithm. Di Blasi et al. \cite{di2005fast} propose a fast photomosaic algorithm to speed up the existing algorithms \cite{silvers1997photomosaics}. Each tile in the database is resized to a $3\times 3$ image. The original image is divided into $3\times 3$ blocks. For each block, the RGB values of the block are used to find the most suitable tile in the tile database, and the found tile is resized to fit the block. Seo et al. \cite{seo2016photomosaic} propose a photomosaic algorithm incorporating social networking activity and relationships between users. The tile database consists of tiles coming from photo albums in the social network of the users. Lee \cite{lee2017automatic} proposes a photomosaic algorithm using adaptive tiling. Unlike other algorithms using tiles with a fixed size, his algorithm adjusts the tile size to adapt to different regions of the original image. Besides these mentioned photomosaic algorithms, there are many other algorithms sharing similar steps (e.g., \cite{lama2014svd, plant2013mosaic}).
\\

Usually, we add a constraint that each tile in the tile database should not be used more than $n_{redu}$ times ($n_{redu}$ is a fixed number denoting the permitted number of redundant tiles for an image) to compose ``nice" photomosaic images. Without the constraint, a tile may be used multiple times in some regions of the photomosaic image, which deteriorates the visual sense of images and conflicts with the goal of photomosaic images (i.e., a photomosaic image should consist of various tiny images). For example, we want to generate a photomosaic image for TV advertisement. Each tiny image includes product information. We want a photomosaic image to include as much product information as possible. Generating a photomosaic image using a limited number of tiny images must be unsuitable. Figure 2 is a photomosaic example without the constraint. We can see that many identical tiles replace the ``sky" region. The photomosaic problem with the constraint is much harder than the same problem without the constraint since the problem becomes a combinatorial optimization problem. Thus the greedy algorithms are not feasible to the problem, and we need new algorithms for the problem.
\\

\begin{figure*}[h!]

\subfigure{
\begin{minipage}{0.48\textwidth}(a)
\centering
\includegraphics[width=5.6cm, height=4cm]{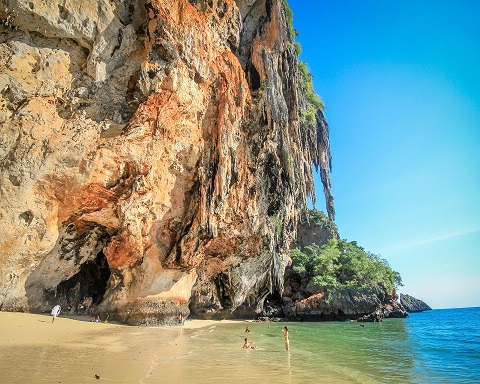}
\end{minipage}
}
\subfigure{
\begin{minipage}{0.48\textwidth}(b)
\centering
\includegraphics[width=5.6cm, height=4cm]{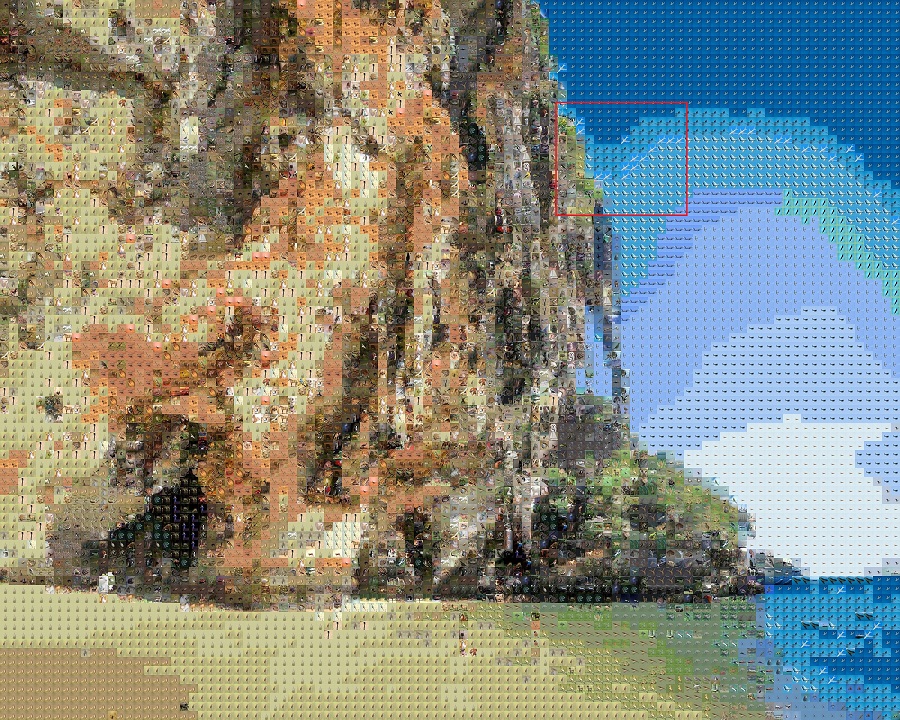}
\end{minipage}
}

\subfigure{
\begin{minipage}{0.48\textwidth}(c)
\centering
\includegraphics[width=5.6cm, height=4cm]{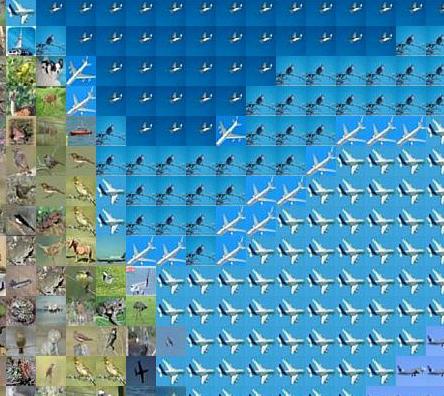}
\end{minipage}
}

\caption{(a) is the original image. (b) is the photomosaic image generated by Mozaika software without the constraint of the number of redundant tiles. (c) is a partition of the photomosaic image. }
\end{figure*}

Evolutionary algorithms (EAs) are a type of optimization algorithms imitating the process of natural selection. They hold the opinion of ``survival of the fittest" \cite{back1997evolutionary}. Coding parameters as chromosomes, EAs search for good solutions by utilizing genetic operations, e.g. crossover, mutation, selection, etc. There are mainly three classes of EAs: evolution strategies (ES), evolutionary programing (EP), and genetic algorithms (GAs) \cite{back1996evolutionary, back2000evolutionary, beyer2013theory}. They have been shown to be good at solving varivous types of optimization problems which other gradient types of mathematical optimizers have failed to solve \cite{man2012genetic}.
\\

Many evidences support that EAs are powerful in combinatorial optimization problems and their applications \cite{wei2015genetic, ray2007genetic, kim2007evolutionary}. However, little work has been done on applying EAs to the photomosaic problem. Sah et al. \cite{sah2008comparison} experimentally investigate the performance of GP and GA on the photomosaic problem. Narasimhan et al. \cite{narasimhan2009randomized} propose a simple randomized local search algorithm called randomized iterative improvement algorithm for the problem. However, these algorithms are time consuming or perform worse than approaches using a greedy algorithm (e.g., \cite{plant2013mosaic, lee2017automatic}). To overcome these drawbacks, we propose a novel evolutionary algorithm called clustering based evolutionary programming (CEP) for the photomosaic problem.
\\

In this paper, a novel clustering based evolutionary programming is proposed to solve the optimization problem. Unlike other EAs, we give the prior knowledge which are the similarities between the tiles to our CEP. Thus CEP is able to converge fast.
\\

The rest of this paper is organized as follows. We formulate the problem in the next section. Section 3 describes our CEP. Section 4 reports the experimental results which indicate CEP outperforms the state of the art algorithms. Section 5 concludes this paper.

\section{Problem formulation}
Suppose we have an original image $im_{1}$. Our goal is to generate a photomosaic image $im_{2}$ which has the minimal mean absolute pixel errors (MAE) to the original image $im_{1}$. Suppose the total number of candidate tiles in the tile database is $n$. Firstly, we divide $im_{1}$ into $n_{r}\times n_{c}$ blocks, and each block consists of $m_{b}\times n_{b}$ pixels. A candidate solution $\overrightarrow{x}=\{x_{1},...,x_{l},...,x_{D}\}$ represents a permutation, where $D=n_{r}\times n_{c}$. Each variable represents the index of a tile for a block. Thus, the variables are discrete, and the range of each variable is 1 to $n$. For example, $x_{1}=100$ means we use the $100^{th}$ tile of the tile database to replace the $1^{st}$ block of $im_{1}$. We add a constraint that each tile in the tile database should not be used more than $n_{redu}$ times. For a fair comparison,  $n_{redu}$ in this paper is the same as \cite{lee2017automatic}.
\\

We normalize the intensities of the blocks and the tiles from [0, 255] to [0, 1]. The cost/fitness function for each block can be formulated as: $fitness(l,k)=|block_{l}-tile_{k}|$,  where $block_{l}$ denotes the $l^{th}$ block of the original image, and $tile_{k}$ denotes the $k^{th}$ candidate tile in the tile database. Since the block and the tile are composed of $m_{b}\times n_{b}$ pixels, the operator $|.|$ is to compute the mean absolute error (MAE) between the pixels of the block and the tile, and we can conclude $0<fitness(l,k)<1$. The overall fitness is computed as: $fitness=\frac{1}{D} \sum_{l=1}^{D} fitness(l,x_{l})$. Our goal is to obtain the solution of the overall fitness function $\overrightarrow{x^{*}}=\mathop{\arg \rm min}\limits_{\overrightarrow{x}}{fitness}$ subject to the constraint that no tile in the tile database is used more than $n_{redu}$ times.
\\

In \cite{narasimhan2009randomized}, an EA is applied to the photomosaic problem. In each evaluation, a randomly selected variable $x_{l}$ mutates to a randomly generated integer $k$ from 1 to $n$ ($x_{l}\leftarrow k$). If $fitness(l,k)<fitness(l,x_{l})$, which denotes the mutation improves the fitness, we accept this mutation. This operation follows the ``survival of the fittest" principle.

\section{Clustering based evolutionary programming}

In this section, we describe CEP. For each tile (tiny image), we generate a normalized color histogram with $B=15$ bins. Then we use K-means algorithm ($K=90$) to cluster the set of normalized histograms. After the clustering, tiles are classified into the different clusters based on their visual similarity. 
\\

The optimization algorithm is composed of the following steps: 1) each block is randomly allocated with a tile, $\overrightarrow{x}$ is used to store the indexes of the tiles for these blocks; 2) we select a random block position $g$ with probability $Prob_{g}= \dfrac{fitness(g,x_{g})}{\sum_{l=1}^{n_{r}\times n_{c}} fitness(l,x_{l})}$; 3) if $fitness(g,x_{g})$ is smaller than the average fitness $\dfrac{1}{n_{r}\times n_{c}} \sum_{l=1}^{n_{r}\times n_{c}} fitness(l,x_{l})$, the current tile $tile_{x_{g}}$ mutates to a randomly selected tile $tile_{k}$ within (or out of) the current cluster with a probability $\alpha$ (or $1-\alpha$) respectively (we choose $\alpha=0.75$ in the paper), while if $fitness(g,x_{g})$ is greater than the average fitness, the probability that the current tile $tile_{x_{g}}$ mutates to a randomly selected tile $tile_{k}$ within (or out of) the current cluster is $1-\alpha$ (or $\alpha$) respectively; 4) if $tile_{k}$ has not been used more than $n_{redu}$ times (i.e., the permitted number of redundant tiles has not been reached), the current tile $tile_{x_{g}}$ mutates to the $k^{th}$ tile $tile_{k}$ ($x_{g}\leftarrow k$), and if the mutation improves the fitness, we accept the mutation. Step 2-4 repeat until a stopping condition is satisfied. The pseudocode of the optimization algorithm is shown in \textbf{Algorithm 1}.
\\

\floatname{algorithm}{Algorithm}  
\renewcommand{\algorithmicrequire}{\textbf{Input:}}  
\renewcommand{\algorithmicensure}{\textbf{Output:}}  
    \begin{algorithm}[!h] 
        \caption{Clustering based Evolutionary Programming }  
        \begin{algorithmic}[1] 
            \Require \\
            $count(\overrightarrow{x},k)$: count the number of $k$ in $\overrightarrow{x}$  \\
            $fitness(x_{1},x_{2})$: compute the MAE between the $x_{1}^{th}$ block and the $x_{2}^{th}$ tile \\
            $n_{r}$: number of tiles for each row\\
            $n_{c}$: number of tiles for each column\\
            $n_{redu}$: permitted number of redundant tiles\\
            $n$: total number of candidate tiles\\
            $\alpha$: a parameter to control the probability that tiles mutate within/out of the cluster they belong to. We set $\alpha=0.75$ in our experiments\\
            \textit{Max\_evaluation}: the maximum number of evaluations
            \Ensure \\
            $\overrightarrow{x}$: the solution for the photomosaic image
            \State
                \State \textbf{initialize} $\overrightarrow{x}$ to be a $n_{r}\times n_{c}$ vector containing a set of random, non-repeating integers from 1 to $n$.
                \While{the total number of evaluations $<\textit{Max\_evaluation}$}
                \State select a random block position $g$ with probability $Prob_{g}= \dfrac{fitness(g,x_{g})}{\sum_{l=1}^{n_{r}\times n_{c}} fitness(l,x_{l})} $.                
                \State generate an uniform, random number $Prob$ between [0, 1].
                \If{$fitness(g,x_{g})< \dfrac{1}{n_{r}\times n_{c}} \sum_{l=1}^{n_{r}\times n_{c}} fitness(l,x_{l})$}
                \State $th\leftarrow \alpha$
                \Else
                \State $th\leftarrow 1-\alpha$
                \EndIf
                \If{$Prob<th$}
                \State generate a random integer $k$ within the cluster that $x_{g}$ belongs to.
                \Else
                \State generate a random integer $k$ out of the cluster that $x_{g}$ belongs to.
                \EndIf
                \If{$count(\overrightarrow{x},k)< n_{redu}$}
                \If{$fitness(g,k)<fitness(g,x_{g})$}
                \State $x_{g}\leftarrow k$
                \EndIf                
                \EndIf
                \EndWhile
                \State \Return $\overrightarrow{x}$
         \end{algorithmic}  
    \end{algorithm}

We design our optimization algorithm based on two assumptions: 1) if a block is similar to a tile (i.e., the MAE between them is small), replacing the current tile by a new tile has a high probability to get a worse result, and if the block is not similar to the tile, the probability should be low. This assumption is intuitive. The total number of candidate tiles for a block is a constant number $n$, thus the number of candidate tiles which are better than the current tile reduces as the fitness (MAE) decreases. Based on this assumption, we design our optimization algorithm such that the tile which is not similar to the block should have a higher chance to mutate into a new tile. This is what step 2 does in the last paragraph. 2) if a block is similar to a tile, we have a high probability to find a more suitable tile within the cluster that the current tile belongs to, and in contrast, if a block is not similar to a tile, the probability should be low. Since visually similar tiles are clustered together, if a tile is visually suitable to a block, the other tiles in the same cluster may also be visually similar to the block. Based on this assumption, we design our optimization algorithm such that the tile which is similar to the block should have a higher chance to mutate within the current cluster and vice versa. This is what step 3 does in the last paragraph.
\\

\begin{figure*}[htbp!]

\subfigure{
\begin{minipage}{0.48\textwidth}(a)
\centering
\includegraphics[width=5.6cm, height=4cm]{original_1.jpg}
\end{minipage}
}
\subfigure{
\begin{minipage}{0.48\textwidth}(b)
\centering
\includegraphics[width=5.6cm, height=4cm]{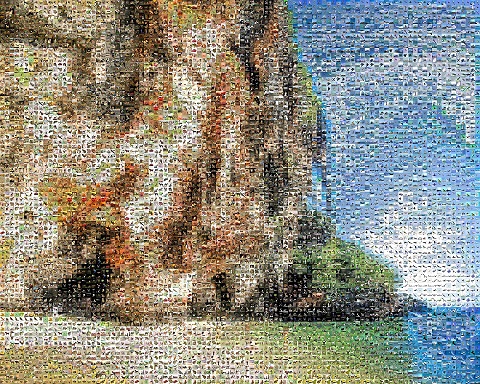}
\end{minipage}
}
\subfigure{
\begin{minipage}{0.48\textwidth}(c)
\centering
\includegraphics[width=5.6cm, height=4cm]{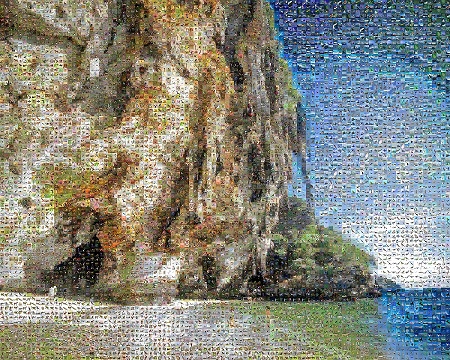}
\end{minipage}
}
\subfigure{
\begin{minipage}{0.48\textwidth}(d)
\centering
\includegraphics[width=5.6cm, height=4cm]{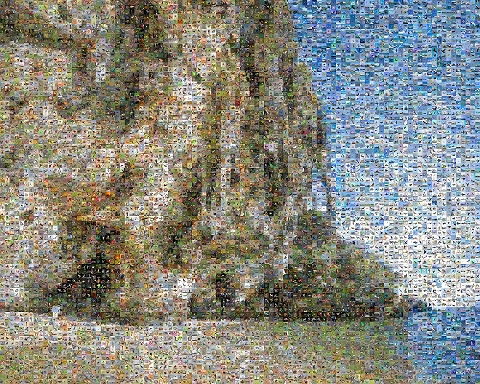}
\end{minipage}
}
\subfigure{
\begin{minipage}{0.48\textwidth}(d)
\centering
\includegraphics[width=5.6cm, height=4cm]{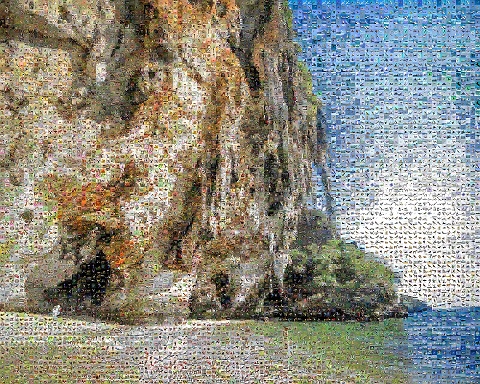}
\end{minipage}
}
\subfigure{
\begin{minipage}{0.48\textwidth}(e)
\centering
\includegraphics[width=5.6cm, height=4cm]{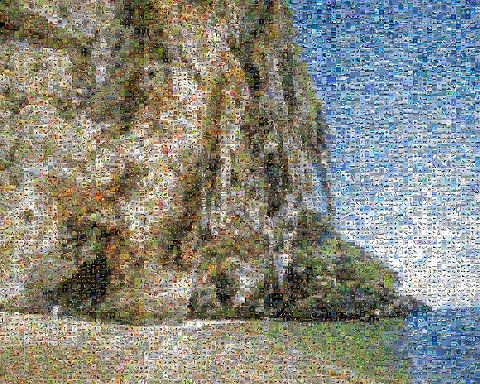}
\end{minipage}
}

\caption{(a) is the original image. (b), (c), (d), (e) and (f) are generated by Andrea software, Mozaika software, Narasimham's algorithm, Lee's algorithm and CEP respectively. }
\end{figure*}

\begin{figure*}[htbp!]

\subfigure{
\begin{minipage}{0.48\textwidth}(a)
\centering
\includegraphics[width=5.6cm, height=4cm]{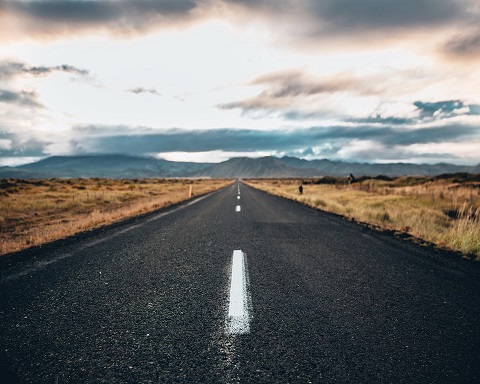}
\end{minipage}
}
\subfigure{
\begin{minipage}{0.48\textwidth}(b)
\centering
\includegraphics[width=5.6cm, height=4cm]{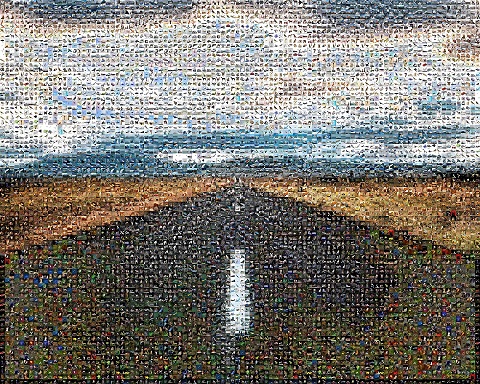}
\end{minipage}
}
\subfigure{
\begin{minipage}{0.48\textwidth}(c)
\centering
\includegraphics[width=5.6cm, height=4cm]{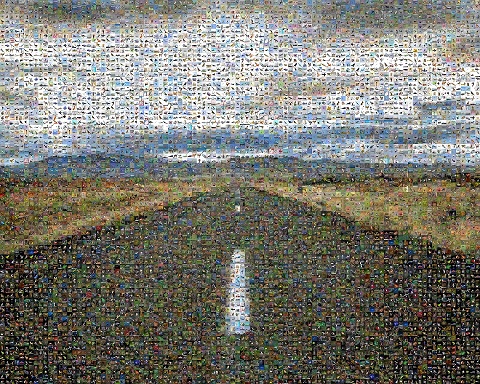}
\end{minipage}
}
\subfigure{
\begin{minipage}{0.48\textwidth}(e)
\centering
\includegraphics[width=5.6cm, height=4cm]{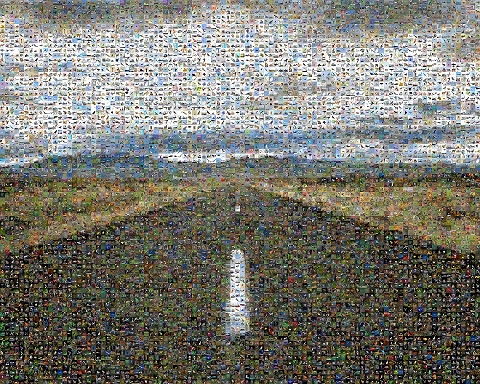}
\end{minipage}
}
\subfigure{
\begin{minipage}{0.48\textwidth}(e)
\centering
\includegraphics[width=5.6cm, height=4cm]{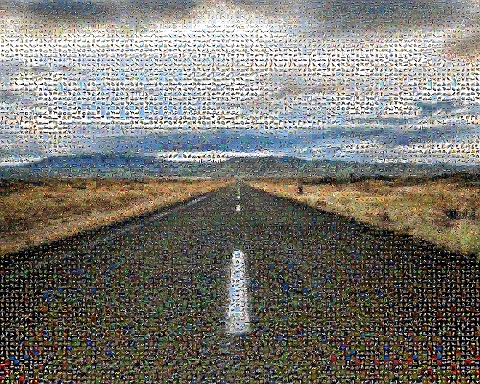}
\end{minipage}
}
\subfigure{
\begin{minipage}{0.48\textwidth}(e)
\centering
\includegraphics[width=5.6cm, height=4cm]{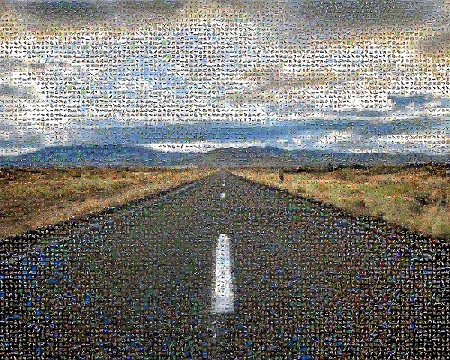}
\end{minipage}
}

\caption{(a) is the original image. (b), (c), (d), (e) and (f) are generated by Andrea software, Mozaika software, Narasimham's algorithm, Lee's algorithm and CEP respectively. }
\end{figure*}

\begin{figure*}[htbp!]

\subfigure{
\begin{minipage}{0.48\textwidth}(a)
\centering
\includegraphics[width=5.6cm, height=4cm]{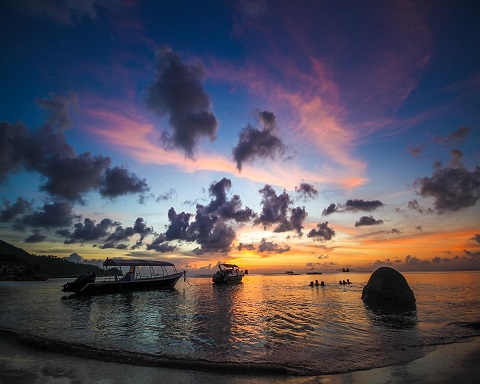}
\end{minipage}
}
\subfigure{
\begin{minipage}{0.48\textwidth}(b)
\centering
\includegraphics[width=5.6cm, height=4cm]{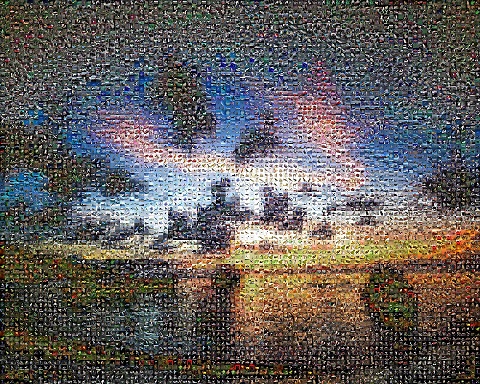}
\end{minipage}
}
\subfigure{
\begin{minipage}{0.48\textwidth}(c)
\centering
\includegraphics[width=5.6cm, height=4cm]{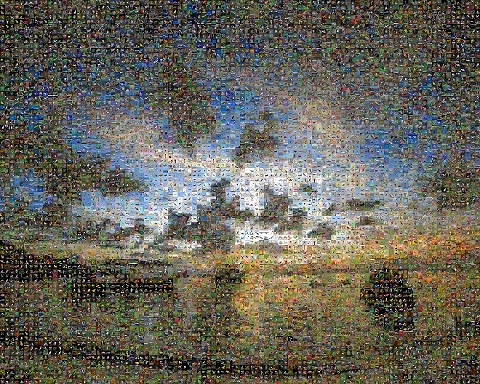}
\end{minipage}
}
\subfigure{
\begin{minipage}{0.48\textwidth}(d)
\centering
\includegraphics[width=5.6cm, height=4cm]{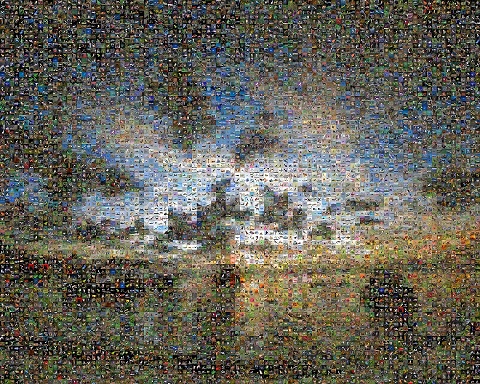}
\end{minipage}
}
\subfigure{
\begin{minipage}{0.48\textwidth}(d)
\centering
\includegraphics[width=5.6cm, height=4cm]{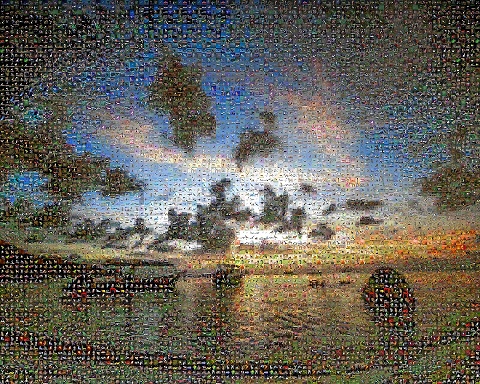}
\end{minipage}
}
\subfigure{
\begin{minipage}{0.48\textwidth}(e)
\centering
\includegraphics[width=5.6cm, height=4cm]{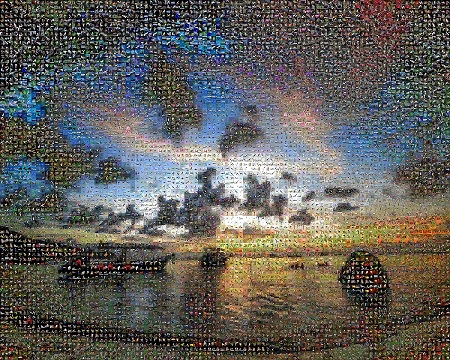}
\end{minipage}
}

\caption{(a) is the original image. (b), (c), (d), (e) and (f) are generated by Andrea software, Mozaika software, Narasimham's algorithm, Lee's algorithm and CEP respectively. }
\end{figure*}

\section{Experimental studies}
To evaluate the performance of the proposed approach, we compare it with Narasimham's algorithm \cite{narasimhan2009randomized}, Lee's algorithm \cite{lee2017automatic}, Andrea software \cite{andrea} and Mozaika software \cite{mozaika}. The latest version of Andrea software is released in January 2018, and the latest version of Mozaika software is released in Feberary 2018. Although Narasimham's algorithm is relatively old (published in 2012), we do not know its efficiency compared with the state of the art algorithms. Thus we investigate the algorithm in our paper. Lee's algorithm is the state of the art algorithm for the photomosaic problem. It can be implemented with adaptive tiling or without adaptive tiling. In Lee's paper, the algorithm without adaptive tiling outperforms the same algorithm with adaptive tiling. Thus we only consider Lee's algorithm without adaptive tiling. Andrea software and Mozaika software are two newest offline software. We use 10000 tiles (tiny images) from imagenet \cite{krizhevsky2009learning} as our tile database. The size of these tiles is $32\times32$. An original image is resized to be a $2560\times3200$ image. Thus the number of tiles for each row is 80 and the number of tiles for each column is 100. Table 1 shows the parameter settings of the following experiments. In the following experiments, the algorithms above are used to generate 30 different photomosaic images \cite{Na} (a few images are shown in figure 3-5 for a qualitative comparison).

\subsection{Qualitative comparison of Photomosaic algorithms}
In this subsection, we qualitatively compare these algorithms. We compare the photomosaic images generated by CEP and Narasimham's algorithm at \textit{Max\_evaluation}=$1.6\times10^{6}$ evaluations (although the performance may still improve as the number of evaluations increases) with the images generated by Lee's algorithm, Andrea software and Mozaika software. We consider computing MAE between a block and a tile is an evaluation. In figure 3-5, photomosaic images (b), (c), (d), (e) and (f) are generated by Andrea software, Mozaika software, Narasimham's algorithm, Lee's algorithm and the proposed algorithm respectively. It is clear that all of the five algorithms are able to generate acceptable photomosaic images.

\subsection{Quantitative comparison of Photomosaic algorithms}
A direct way to evaluate the quality of the visual sense of a photomosaic image is to compare the intensity differences between the photomosaic image and the original image, which is the MAE between them. The performance of CEP and Narasimham's algorithms improves (i.e., the MAE values decrease) as the number of evaluations increases, while for the given parameter settings, Lee's algorithm, Andrea software and Mazaika software have deterministic performance and running time. To fairly compare them, in figure 6, we plot the convergence curves of the average MAE values of CEP and Narasimham's algorithm. The y-axis records the average MAE values, and the x-axis records the number of evaluations. Lee's algorithm, Andrea software and Mazaika software are plotted as flat lines. The y-values of the lines are the average MAE values of them. The three lines are plotted to investigate how many evaluations that CEP and Narasimham's algorithms are needed to outperform the other three algorithms. The figure indicates that only CEP finally outperforms Lee's algorithm.
\\

\begin{table*}[!h]

\caption{Parameter settings of the experiments}
\centering  

\begin{tabular}{p{7cm} p{3cm}   } 
\hline
Image size & $2560\times3200$ \\
Number of candidate tiles& 10000\\
Tile size & $32\times32$		\\
Permitted number of redundant tiles ($n_{redu}$) & 5		\\
Number of tiles for each row & 80		\\
Number of tiles for each column & 100		\\

\hline

\end{tabular}
\end{table*}

\begin{figure*}[h!]

\centering
\includegraphics[width=13cm, height=8cm]{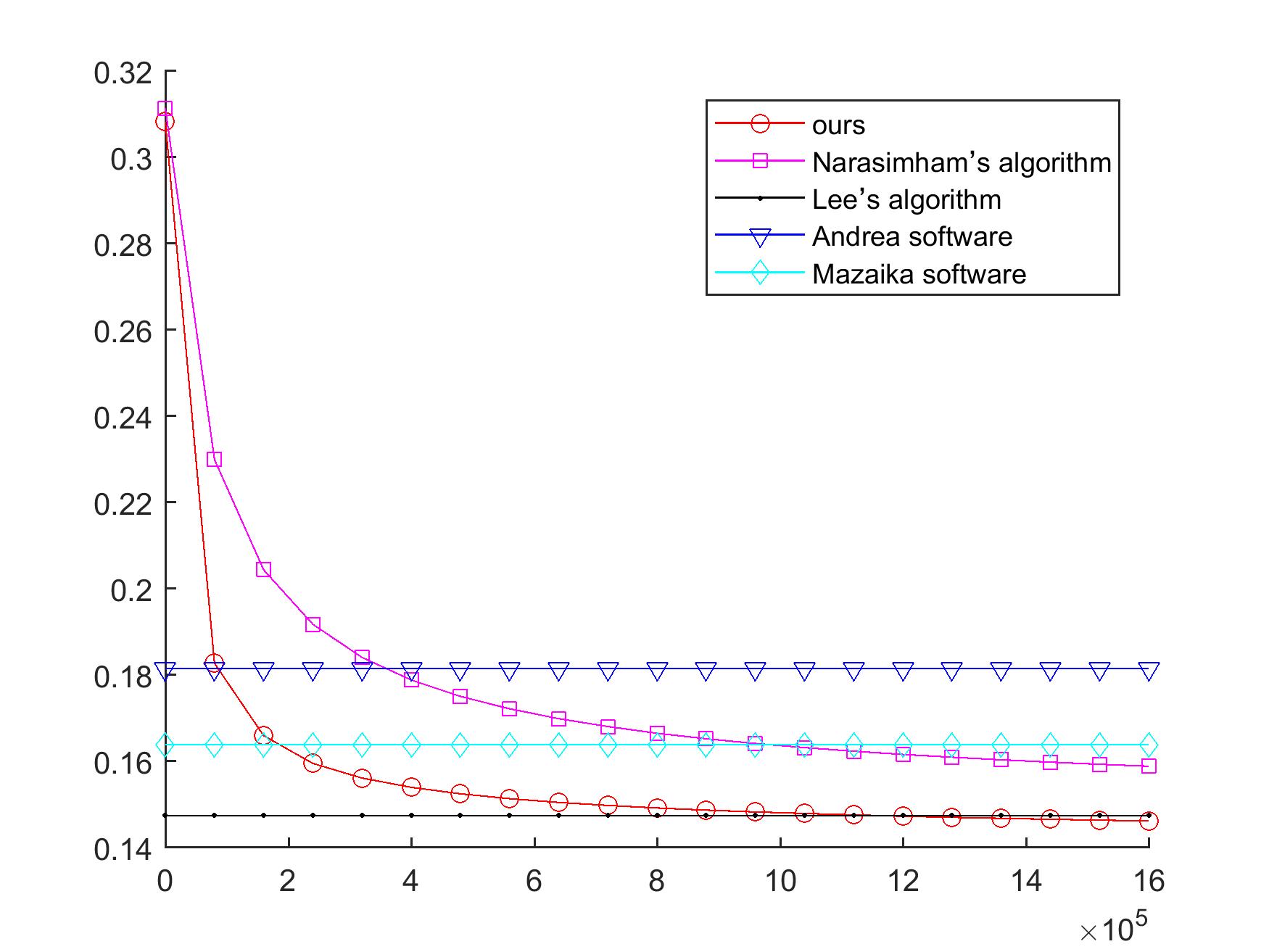}
\caption{The convergence curve of the mentioned algorithms. The y-axis is the average fitness, and the x-axis is the number of evaluations.}
\end{figure*}

Recall that we generate 30 different photomosaic images. Thus we can use Mann-Whitney U-test to test the significant difference between the MAE values of CEP and the other algorithms using a level of significance of 0.05. As we discussed in the previous subsection, we use the MAE values of CEP and Narasimham's algorithm at \textit{Max\_evaluation}=$1.6\times10^{6}$ evaluations. The four algorithms are compared with CEP one by one, and for each comparison, a P-value is obtained and recorded. Table 2 shows the comparison result. Usually, an algorithm is considered to be significantly better if the P-value is smaller than the level of significance. In the table, except Lee's algorithm, the other three algorithms have P-values with CEP smaller than 0.05, which indicates that the proposed algorithm significantly outperforms them.
\\

We implement CEP, Narasimham's algorithm and Lee's algorithm in Matlab platform. However, the algorithms of Andrea software and Mozaika software are not available to the public. Thus we cannot implement them in Matlab platform. Although we download the two software, we cannot fairly compare the running time of the two software with the running time of other algorithms since the two software are compiled using a different programming language. We run CEP, Narasimham's algorithm and Lee's algorithm using a computer which has a CPU with Intel i7-2600 3.40GHz. The running time of these three algorithms are recorded in Table 2. It indicates that the running time of Lee's algorithm is more than three times the running time of CEP. We can conclude that although the visual quality of the photomosaic images generated by CEP and Lee's algorithm has no significantly differences, CEP requires less time to compose a photomosaic image. Narasimham's algorithm is slightly faster than CEP but CEP is able to produce better photomosaic images.

\begin{table*}[!h]

\caption{Comparison result of CEP and the other three algorithms}
\centering  

\begin{tabular}{p{1.8cm} p{1cm} p{1.8cm} p{1.1cm}  p{1.1cm} p{1.1cm} } 
\hline
 & CEP  & Narasimham's & Lee's& Andrea & Mazaika  \\
 Average MAE value &0.1462	&0.1589&	0.1475& 0.1815& 0.1638\\
 Standard deviation & 0.0205&	0.0198&	0.0204&0.0144&0.0214\\
 P-value &NA & 	2.324e-2&0.8418&5.092e-8&3.671e-3 \\
 Average running time &204.05s & 176.12s&632.35s&NA&NA\\

\hline

\end{tabular}
\end{table*}

\section{Conclusion}

Photomosaic images have commercial values. To obtain a better visual sense and satisfy some commercial requirements, a constraint that a tile should not be repeatedly used many times when generating photomosaic images is usually added. With the constraint, composing a photomosaic image becomes a combinatorial optimization problem. EA is a powerful tool to solve combinatorial optimization problems. However, little work has been done on applying EA to the photomosaic problem. In this paper, a clustering based evolutionary programming is proposed to deal with the problem.
\\

The proposed algorithm is qualitatively and quantitatively compared with the state of art algorithms and software. We visually show the photomosaic images of different algorithms. The figures indicate that all algorithm are able to generate acceptable photomosaic images. We evaluate the quality of the generated photomosaic images by comparing the intensity differences between the photomosaic images and the original image. The average fitness convergence curves of these algorithm are plotted indicating CEP performs the best. Moreover, we compare CEP with these algorithms using Mann-Whitney U-test. The comparison result indicates that CEP significantly outperforms these algorithms except Lee's algorithm (which is the state of the art algorithm). However, CEP requires less running time than Lee's algorithm.

\section*{Acknowledgement}
The work described in this paper was supported by a grant from the Research Grants Council of the Hong Kong Special Administrative Region, China [Project No. CityU 125313]. Yaodong He and Jianfeng Zhou acknowledge the Institutional Postgraduate Studentship from City University of Hong Kong. The authors would like to thank Dr.Yang Lou for his valuable comments.

\bibliographystyle{spbasic}

\bibliography{mybib}

\begin{thebibliography}{22}
\providecommand{\natexlab}[1]{#1}
\providecommand{\url}[1]{{#1}}
\providecommand{\urlprefix}{URL }
\expandafter\ifx\csname urlstyle\endcsname\relax
  \providecommand{\doi}[1]{DOI~\discretionary{}{}{}#1}\else
  \providecommand{\doi}{DOI~\discretionary{}{}{}\begingroup
  \urlstyle{rm}\Url}\fi
\providecommand{\eprint}[2][]{\url{#2}}

\bibitem[{Andrea(2018)}]{andrea}
Andrea (2018) http://www.andreaplanet.com/andreamosaic/

\bibitem[{Back(1996)}]{back1996evolutionary}
Back T (1996) Evolutionary algorithms in theory and practice: evolution
  strategies, evolutionary programming, genetic algorithms. Oxford university
  press

\bibitem[{Back et~al(1997)Back, Hammel, and Schwefel}]{back1997evolutionary}
Back T, Hammel U, Schwefel HP (1997) Evolutionary computation: Comments on the
  history and current state. IEEE transactions on Evolutionary Computation
  1(1):3--17

\bibitem[{Back et~al(2000)Back, Fogel, and Michalewicz}]{back2000evolutionary}
Back T, Fogel DB, Michalewicz Z (2000) Evolutionary computation 1: Basic
  algorithms and operators, vol~1. CRC press

\bibitem[{Battiato et~al(2006)Battiato, Di~Blasi, Farinella, and
  Gallo}]{battiato2006survey}
Battiato S, Di~Blasi G, Farinella GM, Gallo G (2006) A survey of digital mosaic
  techniques. In: Eurographics Italian Chapter Conference, pp 129--135

\bibitem[{Battiato et~al(2007)Battiato, Di~Blasi, Farinella, and
  Gallo}]{battiato2007digital}
Battiato S, Di~Blasi G, Farinella GM, Gallo G (2007) Digital mosaic
  frameworks-an overview. In: Computer Graphics Forum, Wiley Online Library,
  vol~26, pp 794--812

\bibitem[{Beyer(2013)}]{beyer2013theory}
Beyer HG (2013) The theory of evolution strategies. Springer Science \&
  Business Media

\bibitem[{Di~Blasi and Petralia(2005)}]{di2005fast}
Di~Blasi G, Petralia M (2005) Fast photomosaic. In: Proceedings of
  International conference in central Europe on computer graphics,
  visualization and computer vision, pp 1--2

\bibitem[{Kim and Choi(2007)}]{kim2007evolutionary}
Kim SJ, Choi MK (2007) Evolutionary algorithms for route selection and rate
  allocation in multirate multicast networks. Applied Intelligence
  26(3):197--215

\bibitem[{Krizhevsky and Hinton(2009)}]{krizhevsky2009learning}
Krizhevsky A, Hinton G (2009) Learning multiple layers of features from tiny
  images

\bibitem[{Lama et~al(2014)Lama, Han, and Kwon}]{lama2014svd}
Lama RK, Han SJ, Kwon GR (2014) Svd based improved secret fragment visible
  mosaic image generation for information hiding. Multimedia tools and
  applications 73(2):873--886

\bibitem[{Lee(2017)}]{lee2017automatic}
Lee HY (2017) Automatic photomosaic algorithm through adaptive tiling and block
  matching. Multimedia Tools and Applications 76(22):24281--24297

\bibitem[{Man et~al(2012)Man, Tang, and Kwong}]{man2012genetic}
Man KF, Tang KS, Kwong S (2012) Genetic algorithms: concepts and designs.
  Springer Science \& Business Media

\bibitem[{Mozaika(2018)}]{mozaika}
Mozaika (2018) http://www.mazaika.com/

\bibitem[{Narasimhan and Satheesh(2009)}]{narasimhan2009randomized}
Narasimhan H, Satheesh S (2009) A randomized iterative improvement algorithm
  for photomosaic generation. In: Nature \& Biologically Inspired Computing,
  2009. NaBIC 2009. World Congress on, IEEE, pp 777--781

\bibitem[{Pictures(2018)}]{Na}
Pictures (2018) https://burst.shopify.com/nature

\bibitem[{Plant et~al(2013)Plant, Lumsden, and Nabney}]{plant2013mosaic}
Plant W, Lumsden J, Nabney IT (2013) The mosaic test: measuring the
  effectiveness of colour-based image retrieval. Multimedia tools and
  applications 64(3):695--716

\bibitem[{Ray et~al(2007)Ray, Bandyopadhyay, and Pal}]{ray2007genetic}
Ray SS, Bandyopadhyay S, Pal SK (2007) Genetic operators for combinatorial
  optimization in tsp and microarray gene ordering. Applied intelligence
  26(3):183--195

\bibitem[{Sah et~al(2008)Sah, Ciesielski, D’Souza, and
  Berry}]{sah2008comparison}
Sah SBM, Ciesielski V, D’Souza D, Berry M (2008) Comparison between genetic
  algorithm and genetic programming performance for photomosaic generation. In:
  Asia-Pacific Conference on Simulated Evolution and Learning, Springer, pp
  259--268

\bibitem[{Seo and Kang(2016)}]{seo2016photomosaic}
Seo S, Kang D (2016) A photomosaic image generation method using photo
  annotation in a social network environment. Multimedia Tools and Applications
  75(20):12831--12841

\bibitem[{Silvers and Hawley(1997)}]{silvers1997photomosaics}
Silvers R, Hawley M (1997) Photomosaics. Henry Holt and Co., Inc.

\bibitem[{Wei et~al(2015)Wei, Tang, and Liu}]{wei2015genetic}
Wei H, Tang XS, Liu H (2015) A genetic algorithm (ga)-based method for the
  combinatorial optimization in contour formation. Applied Intelligence
  43(1):112--131

\end{thebibliography}

\end{document}